\documentclass{article}

\usepackage[preprint]{neurips_2020}

\usepackage[utf8]{inputenc} 
\usepackage[T1]{fontenc}    
\usepackage{hyperref}       
\usepackage{url}            
\usepackage{booktabs}       
\usepackage{amsfonts}       
\usepackage{nicefrac}       
\usepackage{microtype}      
\usepackage{multirow}
\usepackage{eqparbox}
\usepackage{color,soul}

\usepackage{graphicx}
\usepackage[export]{adjustbox}
\usepackage{subcaption}
\usepackage{amsmath, mathrsfs}
\usepackage{mathtools,amssymb,amsthm}

\title{Logit-based Uncertainty Measure in Classification}

\author{%
  Huiyu Wu\thanks{Department of Industrial Engineering and Management Sciences} \\
  \texttt{Northwestern University} \\
  \texttt{huiyuwu2025@u.northwestern.edu} \\
  \And
  Diego Klabjan\footnotemark[1]\\
  \texttt{Northwestern University} \\
  \texttt{d-klabjan@northwestern.edu} \\

}

\begin{document}

\maketitle

\begin{abstract}
  We introduce a new, reliable, and agnostic uncertainty measure for classification tasks called logit uncertainty. It is based on logit outputs of neural networks. We in particular show that this new uncertainty measure yields a superior performance compared to existing uncertainty measures on different tasks, including out of sample detection and finding erroneous predictions. We analyze theoretical foundations of the measure and explore a relationship with high density regions. We also demonstrate how to test uncertainty using intermediate outputs in training of generative adversarial networks. We propose two potential ways to utilize logit-based uncertainty in real world applications, and show that the uncertainty measure outperforms.
\end{abstract}

\section{Introduction}

Machine learning has seen drastic accuracy improvements in classification tasks over the past few years with ever increasing computational power and deeper neural networks. For example, Top-1 Accuracy for ImageNet now exceeds $85\%$ with a state-of-the-art method [1]. Despite the incredible accuracy achievements, neural network classifiers inevitably make mistakes, some of which can be costly. Therefore, it is beneficial to know uncertainty associated with the classification output of neural networks so that we know when a model is more likely to make mistakes.
\begin{figure}[h]
  \centering
  \includegraphics[height=4.5cm]{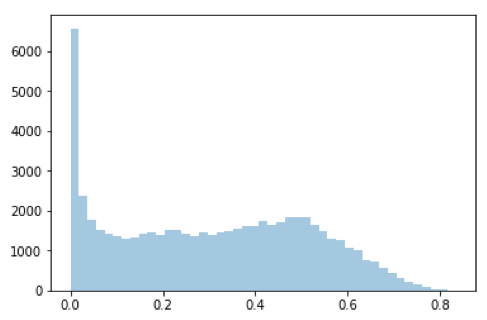}
  \caption{Uncertainty distribution on unrelated data set using ensemble.}
  \label{Intro_uncertainty}
\end{figure}
Many methods of evaluating uncertainty of classification rely upon softmax probabilities generated from neural networks. However, these probabilities or the entropy of these probabilities are notoriously unreliable, making these uncertainty measures unsuitable for tasks such as medical diagnosis or fraud detection when mistakes can be costly. For example, Figure \ref{Intro_uncertainty} shows the uncertainty histogram on FashionMnist that is completely different from the training set MNIST. Uncertainty is obtained using an ensemble method [3] and it shows the high portion of low uncertainty predictions. In this paper, we derive a reliable uncertainty measure based on the logit outputs of neural networks named logit uncertainty. The measure can help classification models detect when they are more likely to make mistakes. The usage of such an uncertainty measure includes, for example, introducing an expert into the decision-making process when the uncertainty associated with classification is high. Another application of such an uncertainty measure is in novelty detection, where a neural network can detect a shift in data distribution [8]. Thus, we can make the decision to retrain the classifier to adapt to the shift. 

\begin{figure}[h]

\begin{subfigure}{0.5\textwidth}
\includegraphics[width=0.9\linewidth, height=4.5cm]{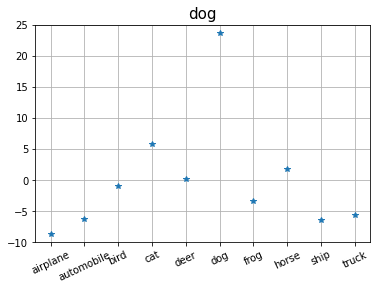} 
\label{Intro_dog}
\end{subfigure}
\begin{subfigure}{0.5\textwidth}
\includegraphics[width=0.9\linewidth, height=4.5cm]{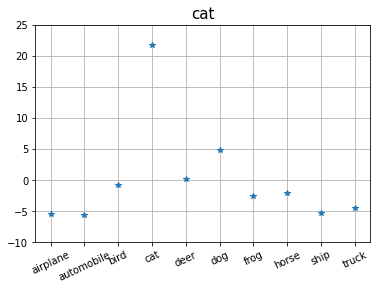}
\label{Intro_cat}
\end{subfigure}

\caption{Average logit values of correctly predicted dog and cat images.}
\label{Intro_dog_cat}
\end{figure}

While logit-based uncertainty applies to any model producing logits such as logistic regression and gradient boosted machines from trees, we focus herein on deep neural networks. Intuitively, logit outputs capture data uncertainty, meaning, if class A is intrinsically similar to class B but different than class C, then the logit value at class B of class A’s logit output is higher than the logit value at class C of the same logit output. For example, in the Cifar10 image classification task [9], we observed that logit values at cat for dog images’ logit outputs are higher compared to the logit values at other classes such as trucks or airplanes and vice versa (Figure \ref{Intro_dog_cat}). Our key idea is to use Gaussian Mixture to model the logit outputs of correctly predicted training sample for each class and to model uncertainty values based on the probability density function of the Gaussian Mixtures. 
The logit-based uncertainty we derive is agnostic and only depends on the raw logit output of the neural network. Therefore, we can treat the neural network classifier as a black box and eliminate the need to retrain the model in order to compute the uncertainty values. This is especially beneficial since training of a complex model on a challenging task can take days or even weeks to complete. As a result, our method to compute uncertainty is easy to incorporate into existing deep learning models in serving.

In the experiments, we demonstrate that logit-based uncertainty outperforms existing uncertainty measures by a large margin. For example, unlike other methods, our uncertainty measure exhibits a clear distinction for images of airplanes, trucks, and cars tested on a neural network trained on passenger cars. Additionally, the order relationship of the uncertainty values obtained from our method is meaningful, with the average uncertainty values for airplanes larger than trucks, and in turn larger than cars for a neural network trained on cars. This is another desired quality that many existing methods do not exhibit [2]. 

Our main contributions are as follows. We propose a new uncertainty measure relying on logit values and agnostic to the underlying classification model. We also analyze theoretical foundations of our method and the relationship between the logit-based uncertainty measure to high density regions. The third contribution is embedded in the comprehensive experimental study. To this end, we introduce an evaluation method consisting of training on one data set and assessing uncertainty on a different data set with context drift. We also experiment by generating GAN [7] images based on a data set and then computing uncertainty on such samples. A comparison of logit-based uncertainty to other prior uncertainty models shows that our strategy is much more reliable. Additionally, we demonstrate that the logit-based uncertainty measure yields better results in two real world applications.

This paper is organized as follows. Section 2 explores existing uncertainty measures. Section 3 introduces and analyzes our method. Section 4 describes the comprehensive experiments. Finally, Section 5 presents conclusions. 

\section{Literature Review}

Our line of work is related to uncertainty and confidence estimation for deep learning classification, which aims to generate a number typically within $0$ and $1$ from outputs of neural networks. Sensoy et al. [5] provide a novel idea of viewing the logit outputs of neural networks as evidence for each class, and build a Dirichlet distribution with parameters based on those evidences. The authors cleverly tailor a loss function that suits their need and are able to generate meaningful uncertainty measures. However, in practice, their method requires retraining of the neural network which can be expensive. Additionally, their uncertainty values tend to concentrate around 1 when encountering novel data points, making the order relationship of the uncertainty values less meaningful. Our method addresses both of the problems and achieves stronger results in more general tests.

Many other existing methods that compute uncertainty require a number of outputs for a single data sample. These methods typically generate the uncertainty values base on the average or the variance of the outputs. These include the ensemble method [3], the drop-out method [4], and the Bayesian neural networks [6]. Similar to logit-based uncertainty values, their values do not concentrate on a single value compared to [5] and offer some evidence of a meaningful order relationship with higher uncertainty actually representing lower confidence. However, from the experiments, when encountering out-of-the-distribution data points, these methods exhibit poor performance compared to the logit-based uncertainty method. Additionally, since all of the above require multiple outputs for a single data sample, they all require additional computational power, which is often not practical.

One method to compute uncertainty is provided in [2]. The authors propose to first construct an $\alpha$-high-density-set of each class and then compute the uncertainty based on the distances of the given points to the constructed sets. Similar to logit-based uncertainty, this uncertainty measure is agnostic to the underlying model and only requires minimal computational resources to generate the uncertainty value. However, this method does not work well on high-dimensional data sets such as MNIST, Cifar10, and Cifar100. Logit-based uncertainty does not construct density sets of each class, but models the logit outputs using mixtures directly and it empirically obtains much better uncertainty estimates on these high-dimensional data sets.

Another uncertainty method is introduced in [11] named deep k-NN (DkNN). This method compares the intermediate output of deep neural networks of training and testing sets, and it uses a nearest neighbors method to estimate nonconformity in the predictions. The nonconformity can then be used to make predictions, confidence, and reliability. This method has very reliable performance but requires extensive computational power. For example, in our experiments on finding uncertainty of GAN images, DkNN requires about $19$ hours computating on Tesla P100 GPU, and for comparison, our method only requires less than $5$ minutes on the same GPU.

\section{Methodology}

We have observed that logit values capture some uncertainties, meaning that the logit outputs of a specific class should share similar logit values in each dimension of the logit vector. The intuitive explanation for the above logic is that the samples from one class should share similar features. Therefore, if the logit values of a sample is different from the known logits of predicted class, we expect the prediction to be of low confidence. With such intuition in mind, we work with the raw logit outputs of the correctly predicted training data. 

In summary, we fit Gaussian Mixture Models (GMMs) to the logit vectors of correctly classified samples for each class; reasons for using GMMs are they are universal approximations [12], and under certain assumptions, a neural network generates GMM distributed outputs. Details are provided in Section 3.2. One natural idea that follows is to base the uncertainty measure on the density functions of the GMMs; meaning the larger the density function value the smaller the uncertainty. To work with extreme small values of the density functions that are typically observed in datasets, we design a score function to make the values more manageable. Lastly, we use a Sigmoid function to map the values obtained from the score function to the range between $0$ and $1$. 

\subsection{Uncertainty Measure}

Suppose we have a data set with $k$ classes and we have already trained a classification model on this data set. For each class $i$, we select the correctly predicted training data and use a GMM to model the logit outputs of these data. The number of components is selected based on the Bayesian Information Criterion against the number of components and we use the elbow rule to select a reasonable number of components. Suppose the fitted Gaussian Mixture has probability density function $gmm_i$. We then define score $s_i(X)$ for a feature vector $X$ as
\begin{equation}
    s_i(X) = \ln(\max_t(gmm_i(t))) - \ln(gmm_i(X)).
\end{equation}

If $X$ is considered as a random vector, then the score $s_i$ is also a random vector, and we can find the $q$-th percentile of the score denoted as $s_{iq}$.To map score $s$ to $[0, 1]$ we apply the logistic function $g_i(s) = \frac{1}{1 + e^{-c_{i1}(s-c_{i2})}}$.
In order to find the parameters $c_{i1}$ and $c_{i2}$ of the logistic function, we use four hyperparameters $0\leq u_{1}\leq1, 0\leq u_{2}\leq1, 0\leq q_{1}\leq1,$ and $0\leq q_{2}\leq1$. We want the $q_{1}$ quantile of $s_i$ to map to $u_{1}$ and $q_{2}$ to $u_{2}$. This translates to $g_i(s_{iq_{1}}) = u_{1}$, and $g_i(s_{iq_{2}}) = u_{2}$, resulting in
\begin{equation}\label{constants}
    \begin{cases}
    c_{i2} = \frac{s_{iq_{2}}\ln(u_{1}^{-1} - 1) - s_{iq_{1}}\ln(u_{2}^{-1} - 1)}{\ln(u_{1}^{-1} - 1) - \ln(u_{2}^{-1} - 1)}\\

    c_{i1} = \frac{-\ln(u_{2}^{-1} - 1)}{s_{iq_{2}} - c_{i2}}.\\
    \end{cases}
\end{equation}
Finally, when we encounter a new data sample $x$ that is classified as class $i$, its uncertainty value is: $u(x) = g_i(s_i(x))$. 

\subsection{Analysis}

In this section, we first provide reasons for selecting Gaussian Mixtures to model the logit output vectors for each class. Then we analyze the proposed logit uncertainty to make sure it makes intuitive sense. Additionally, we explore the relationship of the logit-based uncertainty measure with existing statistical concepts. 

We start with reasons for using Gaussian Mixtures. We first describe a theorem that shows with proper assumptions that the limiting distribution of neural network outputs converges in distribution to Gaussian Mixture. Consider a fully connected neural network with $D$ hidden layers each of size $H_\mu$, $\mu \in \{1, \dots, D\}$. The network has real-valued input and output vectors of dimensions $H_0$ and $H_{D+1}$, respectively. We describe the neural network with the following recursion [13]:

\begin{equation}
    f^{(1)}_{i}(x) = \sum_{j=1}^{M}w_{ij}^{(1)}x_j+b_i^{(1)},
\end{equation}
\begin{equation}
    g^{(\mu)}_{i}(x) = ReLU(f^{(\mu)}_{i}(x)),
\end{equation}
\begin{equation}
    f^{(\mu+1)}_{i}(x) = \sum_{j=1}^{H_{\mu}}w_{ij}^{(\mu+1)}g^{(\mu)}_{j}(x)+b_i^{(\mu+1)},
\end{equation}

where $i$ ranges between $1$ and $H_{\mu}$. We also assume 

\begin{equation}
    w_{ij}^{(\mu)}\sim\mathcal{N}(0,C_{w}^{(\mu)})\hspace{3mm}  i.i.d., \mu \in \{1, \dots D+1\},
\end{equation}
\begin{equation}
    b_{i}^{(\mu)}\sim\mathcal{N}(0, C_b^{(\mu)})\hspace{3mm}   i.i.d., \mu \in \{1, \dots D\},
\end{equation}
\begin{equation}
    b_i^{D+1}\sim G(c)\hspace{3mm}   i.i.d.,
\end{equation}

Here $G(c)$ denotes a Gaussian Mixture distribution with $c$ components. We have a sequence of neural networks $n = 1, 2, \dots$ with network $n$ having $H_\mu(n)$ neurons in layer $\mu$ for $\mu = 1,\dots,D$. The number of layers $D$ does not depend on $n$. Additionally, since we are interested in the limiting behavior of the neural network, we scale the weight variances to ensure a converging variance where $\hat{C}_w^{(\mu)} = C_w^{(\mu)}H_{\mu-1}(n) < Q$ for every $\mu$ and $n$. We now present the result.

\paragraph{Theorem 1.} Consider a random fully connected neural network of the form (3)-(5) satisfying assumptions (6)-(9). Then for all sets of strictly increasing in $n$ width functions $H_\mu(n)$ and for any finite input set $\{x_m\}_{m=1}^{K}$, the distribution of the output neural network converges in distribution to a Gaussian Mixture distribution as $n \to \infty$.

The proof of this theorem is in the Appendix. The next claim ensures logit-based uncertainty makes intuitive sense. The uncertainty we are trying to compute is similar to the problem of estimating the probability that a new data point in the k-dimensional Euclidean space belongs to the Gaussian mixture for the predicted class. It makes sense that points with higher density values should have lower uncertainties.

\paragraph{Proposition 1.} For $x_1$ and $x_2$ that are predicted as class $i$, if $gmm_i(x_1) > gmm_i(x_2)$, then we have $u(x_1) < u(x_2)$. 

This proposition follows from monotonicity. Before moving onto the second claim, we introduce the highest density region (HDR) [10]. The $(1-\alpha)$-HDR is the subset $R(f_\alpha)$ of sample space $X$ such that $R(f_\alpha) = \{x: f(x) \geq f_\alpha\}$, where $f_\alpha$ is the largest constant such that $P(X \in R(f_\alpha)) \geq 1 - \alpha$ with $f(x)$ being the probability density function of $X$. We have the following result that builds a connection between logit-based uncertainty and HDR.

\paragraph{Proposition 2.} Any sample $x$ within the $q_1$-HDR has uncertainty value $u(x)< u_1$. Similarly, any sample $x$ within the $q_2$-HDR has uncertainty $u(x)< u_2$.

The second proposition builds a connection between the proposed method of logit-based uncertainty and HDR. The claim also enables us to adjust the parameters of logit-based uncertainty based on different needs. For example, if we are detecting tumors and misclassifications are costly, we can adjust the model to be more conservative with lower values of $q_1, u_1, q_2, u_2$.

Another result follows from the second claim. For confidence value $\kappa \in [0, 1]$ and confidence region $R \subseteq \Omega$ for a probability density function, $\kappa$ is a confidence value related to $R$ if $\int_{\Omega \textbackslash R} p(x) dx = \kappa$.

We note that the confidence value is not useful unless $R$ is defined as a minimal volume region that satisfy the integral. The resulting confidence region $R$ with confidence value $\kappa$ then becomes the HDR $R(f_\kappa)$ [10]. Therefore, we can rephrase proposition 2 as any sample $x$ within the $(1 - q_1)$ confidence region has uncertainty value $u(x)< u_1$ and any sample $x$ within the $(1 - q_2)$ confidence region has uncertainty value $u(x)< u_2$.

For Gaussian mixtures with one component, the multivariate Gaussian case, we explore the relationship between the logit uncertainty and the Mahanobis distance to Gaussian distributions. The Mahanobis distance $r(x)$ between $x$ that is sampled from a Gaussian and the same Gaussian distribution with mean $\mu$ and covariance matrix $\Sigma$ is defined as $r(x)= ((x - \mu)^T\Sigma^{-1}(x - \mu))^{1/2}$. Furthermore, let $F(r)$ be the cumulative distribution function of random vector $r$. The following result builds a connection between HDR and $r$.

\paragraph{Proposition 3.} For $(1 - \alpha)$-HDR, let $R(f_\alpha)$ be defined with respect to a multivariate Gaussian distribution with mean $\mu$ and covariance matrix $\Sigma$, and probability density function $\phi$. If we let $r_\alpha = F^{-1}(1 - \alpha)$, then for any $x \in R(f_\alpha)$ the Mahanobis distance with respect to the Gaussian distribution satisfies $r(x)< r_\alpha$.

Combining Propositions 2 and 3, it follows that when the Gaussian Mixture has only one component, samples that share the same logit-based uncertainty have the same Mahanobis distance to the Gaussian distribution.

\subsection{Extensions}

Logit-based uncertainty is not applicable to k-nearest neighbors, random forests, or decision trees. These methods, when used for classification, do not output values that can be fitted using a GMM, therefore, our method does not apply to these classifiers. However, logit-based uncertainty applies to logistic regression, support vector machines, and gradient boosting machines. For logistic regression, it can be viewed as a single layer neural network therefore our logit uncertainty applies. Linear SVMs use a hyperplane to separate different classes and output the distance to the hyperplane given a data point. To obtain logit-based uncertainty, we can fit GMMs to the distances obtained from the training data set. Therefore, our method also applies to SVMs. A gradient boosting machine outputs logits for each class when used for classification which makes it easy to compute logit-based uncertainty. 

\section{Experiments}

In this section we empirically evaluate the proposed approach, assessing the performance of logit uncertainty on different tasks described below. In all of the below tests, we use $q_1 = 80$ and $q_2  = 60$ with $u_1  = 0.5$ and $u_2 = 0.2$, as these hyper-parameters yield the most robust results across different experiments. The experiments are implemented using Tensorflow and all training and GAN experiments are done on Google Cloud with Tesla P100 GPU and the remaining experiments are timed on Google Colab with Tesla T4 GPU.

\subsection{MNIST Experiments}

\begin{figure}[t]

\begin{subfigure}{0.5\textwidth}
\includegraphics[width=0.9\linewidth, height=4.5cm]{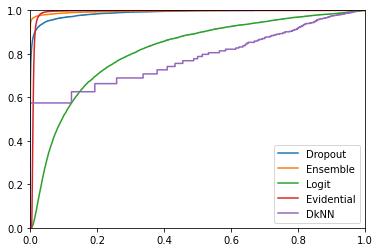} 
\caption{Correct predictions}
\label{mnist_correct}
\end{subfigure}
\begin{subfigure}{0.5\textwidth}
\includegraphics[width=0.9\linewidth, height=4.5cm]{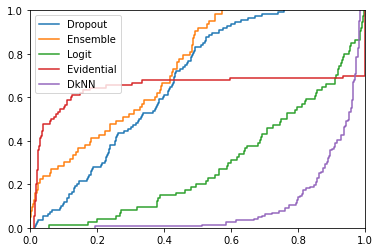}
\caption{Incorrect predictions}
\label{mnist_incorrect}
\end{subfigure}

\caption{Uncertainty distribution for correct and incorrect predictions on MNIST test set.}
\label{mnist_test}
\end{figure}

MNIST is a data set of handwritten images with 60,000 training and 10,000 testing samples, where each image is of size 28 $\times$ 28. For training of MNIST, we use a CNN with 20 and 50 filters with size 5 $\times$ 5 and 500 hidden units for the fully connected layer. This simple architecture achieves performance similar to state-of-the-art. The first test involves inspecting the uncertainty distribution on the correctly and incorrectly classified images in the test set; Figure \ref{mnist_test} shows the empirical CDF of uncertainty values for such images. As a comparison, we also include the uncertainty distributions using the ensemble method detailed in [3], Bayesian approximation from [4], the Evidential method from [5], and the deep k-NN method from [11]. It is worth noting that classification accuracy is about $99\%$ for all five methods.
From the above results, DkNN demonstrates a better performance, however, as mentioned in Section 2, DkNN requires many more computational resources.  Compared to other methods, logit-based uncertainty shows better performance on incorrect predictions.

\begin{figure}[h]
  \centering
  \includegraphics[height=4.5cm]{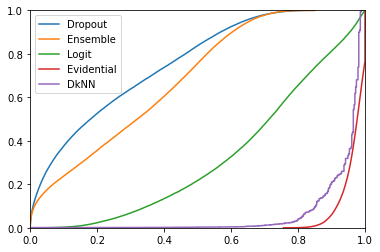}
  \caption{Uncertainty distribution for FashionMNIST.}
  \label{fashion}
\end{figure}

Another experiment involves context drift. We use our model trained on the MNIST data set and then perform inference on the FashionMNIST data set. FashionMNIST is a data set of fashion images including clothes, dresses, and shoes, with each image of size 28 $\times$ 28. We should expect a perfect uncertainty measure to output values close to $1$ as uncertainty for all predictions since we are forcing the model to classify fashion items as digits. Figure \ref{fashion} shows the empirical CDF of uncertainty values for 60,000 training FashionMNIST images using logit-based uncertainty method and other benchmark methods.
Although the Evidential method seems to have the best performance with the majority of uncertainty values above $0.8$, we are going to, in the following experiments, observe that the Evidential method tends to be over-conservative. DkNN also shows superior performance, but again, it requires much more computing power. Compared to Ensemble and Dropout, logit-based uncertainty exhibits preferred behavior in that it does not have low uncertainty values, and has many more predictions with high uncertainty values. 

\begin{figure}[h]
  \centering
  \includegraphics[height=4.5cm]{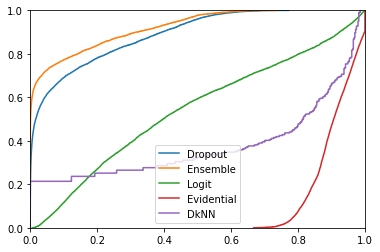}
  \caption{Uncertainty distribution for USPS data set.}
  \label{USPS}
\end{figure}

We next introduce a new experiment regarding uncertainty measures by testing them on a data set where we should expect a perfect uncertainty measure to output intermediate values. To be more specific, for a model trained on MNIST, the handwritten digits data set, we expect the model to output intermediate values for fuzzy handwritten images. To this end, we select the USPS data set. The USPS data set consists of 7,291 training images of handwritten digits, however, the crucial difference between the USPS and MNIST is that the size of each USPS image is 16 $\times$ 16 whereas the size of each MNIST image is 28 $\times$ 28. Therefore, if we enlarge the USPS image to 28 $\times$ 28, we expect a perfect uncertainty measure to produce intermediate uncertainty values. Figure \ref{USPS} shows the uncertainty empirical CDF of the models trained on MNIST and tested with enlarged USPS images. 
Comparing to the results obtained in Figures \ref{mnist_test} and \ref{fashion} it is obvious that for the USPS test, the logit-based uncertainty measure outputs many more uncertainty values from the $0.2$ to $0.8$ range, which is what we expect for a good uncertainty measure. Other methods have more extreme uncertainty values, and as we have anticipated, Evidential tends to be very conservative and prefers to label predictions with high uncertainty values. DkNN shows a similar performance to logit-based uncertainty.

\begin{figure}[h]
  \centering
  \includegraphics[height=4.5cm]{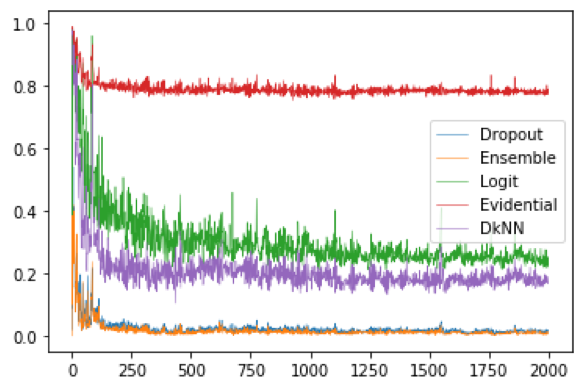}
  \caption{Uncertainty distribution for GAN images.}
  \label{GAN}
\end{figure}

Similar to the previous test, we create another test to establish if logit-based uncertainty values are meaningful. We expect that as training of GAN proceeds [7], the generator’s images are approaching real images of the target data set. Therefore, we use a generator with 64 and 128 filters of size 5 $\times$ 5 at the first and second convolution layers to generate images of digits similar to that from the MNIST data set. We trained the GAN for 2,000 epochs and let each epoch generate 256 images using the generator. We then use the CNN model trained on MNIST to classify these images and compute their logit uncertainty. We only show the result using GAN to generate images of handwritten digit 7 since the results for other digits are similar. Figure \ref{GAN} shows the average of the 256 uncertainty values after each of the 2,000 training epochs. 
We know that during GAN training, generator’s images are improving from fuzzy to clear as the training epochs increase, thus we should expect a perfect uncertainty measure to have uncertainty values for these images change from high uncertainty at the start of training to low uncertainty at the end of training. Indeed, this is what we observe in Figure \ref{GAN} with logit-based uncertainty. From the same plot, we observe that the Ensemble and Dropout uncertainty distributions share very different shapes with the logit uncertainty distribution. These two methods output relatively low uncertainty when the training epoch number is low, which should not be the case for a good uncertainty measure. Evidential, given its conservative tendency, indeed has high uncertainty values even after 2,000 epochs of training. Again, DkNN shows a similar performance to logit-based uncertainty.

\begin{table}[h!]
\centering
\caption{Uncertainty Statistics}
\begin{tabular}{ |c|c|c|c|c|c|c| } 
 \hline
 \multicolumn{2}{|c|}{Test and statistics} & Logit & Dropout & Ensemble & DkNN & Evidential \\ \hline
 \multirow{ 2}{*}{MNIST Correct} & Mean & \eqmakebox[data][r]{\textbf{0.185}} & \eqmakebox[data][r]{0.011} & \eqmakebox[long][r]{0.004} & \eqmakebox[data][r]{0.219} & \eqmakebox[long][r]{0.011}\\ 
 & Skewness & \eqmakebox[data][r]{1.819} & \eqmakebox[data][r]{6.780} & \eqmakebox[long][r]{10.944} & \eqmakebox[long][r]{\textbf{1.073}} & \eqmakebox[long][r]{33.663}\\ \hline
 \multirow{ 2}{*}{MNIST Incorrect} & Mean & \eqmakebox[data][r]{0.714} & \eqmakebox[data][r]{0.328} & \eqmakebox[long][r]{0.264} & \eqmakebox[data][r]{\textbf{0.902}} & \eqmakebox[long][r]{0.355} \\
 & Skewness & \eqmakebox[data][r]{\textbf{-0.722}} & \eqmakebox[data][r]{0.156} & \eqmakebox[long][r]{-0.050} & \eqmakebox[data][r]{-2.801} & \eqmakebox[long][r]{0.737}\\ \hline
 \multirow{ 2}{*}{FashionMNIST} & Mean & \eqmakebox[data][r]{0.680} & \eqmakebox[data][r]{0.241} & \eqmakebox[long][r]{0.314} & \eqmakebox[data][r]{\textbf{0.939}} & \eqmakebox[long][r]{0.967} \\
 & Skewness & \eqmakebox[data][r]{\textbf{-0.551}} & \eqmakebox[data][r]{0.637} & \eqmakebox[long][r]{0.073} & \eqmakebox[data][r]{-4.463} & \eqmakebox[long][r]{-1.379}\\ \hline
 \multirow{ 2}{*}{USPS} & Mean & \eqmakebox[data][r]{\textbf{0.457}} & \eqmakebox[data][r]{0.109} & \eqmakebox[long][r]{0.077} & \eqmakebox[data][r]{0.622} & \eqmakebox[long][r]{0.905} \\
 & Skewness & \eqmakebox[data][r]{\textbf{0.383}} & \eqmakebox[data][r]{1.561} & \eqmakebox[long][r]{2.028} & \eqmakebox[data][r]{-0.736} & \eqmakebox[long][r]{-0.383}\\ \hline
\end{tabular}
\end{table}

Table 1 provides some summary statistics. From the table we see that logit-based and DkNN uncertainty measures achieve good performance. They produce low uncertainty values for correctly predicted MNIST, high uncertainty values for incorrectly predicted MNIST and FashionMNIST, and modest uncertainty values for USPS. Skewness is an informative statistics as well. For correctly predicted MNIST we prefer a small positive skewness, and for incorrectly predicted MNIST and FashionMNIST, we prefer a negative skew with small absolute value. For USPS we prefer to have a small skewness in absolute value. We observe that logit-based uncertainty and DkNN meet all of the expectations whereas other methods all have at least one or more statistics that do not meet expectations. Therefore, for each statistics we compare and highlight the better between logit-based uncertainty and DkNN. The two methods achieve similar performance, however, in the next section we compare the running time and memory usage between the two methods, and argue that considering computing efficiency, logit-based uncertainty is better suited for the different tasks.  

\subsection{Running Time and Memory}
Dropout and ensemble do not require long running time and much memory with an already trained model. Evidential method requires retraining since it utilize a special loss function, however, with a trained model, evidential does not require much computing resource to compute uncertainty. Therefore, in this section, we only compare the running time and memory usage of logit-based uncertainty and DkNN, with Table 2 summarizing the results. 
\begin{table}[h!]
\centering
\caption{Running Time and Memory Comparison}
\begin{tabular}{ |c|c|c| } 
 \hline
 Memory & Logit & DkNN/Logit \\ \hline
 Setup & 0.58mb & 10.7\\ \hline \noalign{\vskip 2mm} \hline 
 Time & Logit & DkNN/Logit\\ \hline
 Setup & \eqmakebox[long][r]{4.1s} & \eqmakebox[long][r]{113.9}\\ 
 MNIST test set & \eqmakebox[long][r]{14.8s} & \eqmakebox[long][r]{38.2}\\ 
 FashionMNIST & \eqmakebox[long][r]{12.6s} & \eqmakebox[long][r]{46.5}\\ 
 USPS & \eqmakebox[long][r]{10.5s} & \eqmakebox[long][r]{51.7}\\ 
 GAN images & \eqmakebox[long][r]{300.0s} & \eqmakebox[long][r]{228.0}\\ \hline 
\end{tabular}
\end{table}
 From the comparison we see that the running time for logit-based uncertainty is at least about 40 times faster than DkNN. Logit-based uncertainty also scales better since most of the computing time is to compute model output, and if we are starting with logit outputs, it needs only little additional computation. In the GAN experiments with roughly half a million images, logit-based uncertainty is more than 200 times faster than DkNN. Additionally, for the MNIST test, DkNN requires more than 10 times the memory used by logit-based uncertainty. The memory difference would get worse with more complicated neural networks since DkNN requires the storage of all intermediate layer outputs. We conclude that logit-based uncertainty is a more well rounded method compare to DkNN.

\subsection{Cifar10 Experiments}

\begin{figure}[t]

\begin{subfigure}{0.5\textwidth}
\includegraphics[width=0.9\linewidth, height=4.5cm]{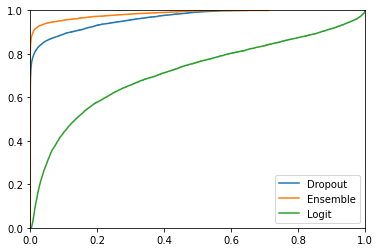} 
\caption{Correct predictions}
\label{cifar10_correct}
\end{subfigure}
\begin{subfigure}{0.5\textwidth}
\includegraphics[width=0.9\linewidth, height=4.5cm]{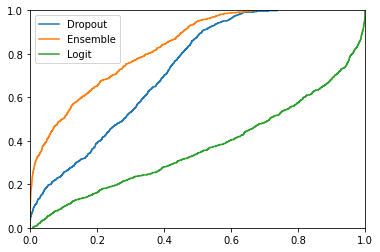}
\caption{Incorrect predictions}
\label{cifar10_incorrect}
\end{subfigure}

\caption{Uncertainty distribution for correct and incorrect predictions on Cifar10 test set.}
\label{Cifar10_test}
\end{figure}

The Cifar10 data set consists of 60,000 color images of ten object classes with each image of size 32 $\times$ 32. We use Densenet40 with $k = 12$ for training of Cifar10 [9] and then perform similar tests used for the MNIST data set. In the following tests, we do not include evidential and DkNN. The implementation of the Evidential method is not stable and does not achieve meaningful accuracy. DkNN is excluded due to memory limitations, as it requires storage of all the intermediate outputs. Accuracies for the remaining models are comparable, and are about $92\%$. In the first test, we examine the empirical CDF of uncertainty values of correctly and incorrectly classified images of the test set (Figure \ref{Cifar10_test}). 

\begin{figure}[t]

\begin{subfigure}{0.5\textwidth}
\includegraphics[width=0.9\linewidth, height=4.5cm]{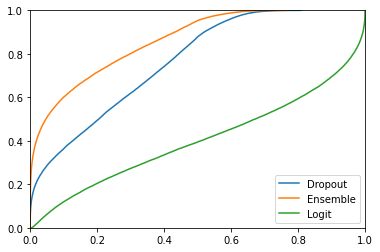} 
\caption{Uncertainty on Cifar100}
\label{cifar100}
\end{subfigure}
\begin{subfigure}{0.5\textwidth}
\includegraphics[width=0.9\linewidth, height=4.5cm]{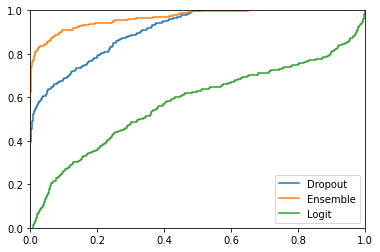}
\caption{Uncertainty for pickup trucks predicted as Auto}
\label{pickup}
\end{subfigure}

\caption{Uncertainty distribution on unseen data set.}
\label{cifar10_additional}
\end{figure}

The second test involves testing the model trained on Cifar10 on a different data set. For this test, we select the Cifar100 data set, consisting of 60,000 color images of 100 classes, where each image is 32 $\times$ 32 in size. The 10 classes of Cifar10 are different from the 100 classes in the Cifar100 data set. Therefore, we should expect a good uncertainty measure to output high uncertainty values, which is observed for the logit-based uncertainty in the empirical CDF in Figure \ref{cifar100}.

The third test for Cifar10 is on images that we should expect a good uncertainty measure to output intermediate values. To achieve this goal, we observe that one of the Cifar10 classes is automobile with images of passenger cars that do not include pickup trucks. Additionally, Cifar100 has a pickup truck class and, since pickup trucks and passenger cars are intrinsically similar to each other, we expect the uncertainty values of pickup trucks that are predicted as automobiles to be larger than those of the automobile images in the Cifar10 test set yet smaller than those of Cifar100. This is exactly what we observe in Figure \ref{pickup} for logit-based uncertainty.

\subsection{Uncertainty and Embeddings}

\begin{figure}[t]

\begin{subfigure}{0.5\textwidth}
\includegraphics[width=0.9\linewidth, height=4.5cm]{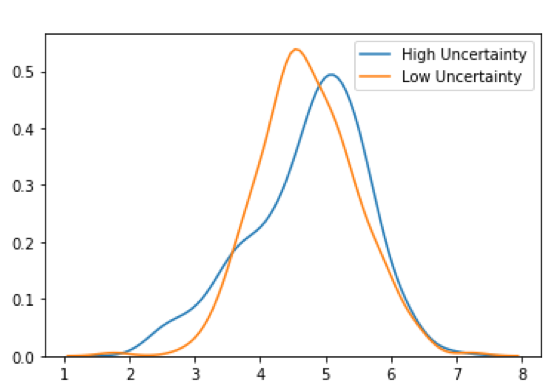} 
\caption{Minimum distance distribution.}
\label{min_dist}
\end{subfigure}
\begin{subfigure}{0.5\textwidth}
\includegraphics[width=0.9\linewidth, height=4.5cm]{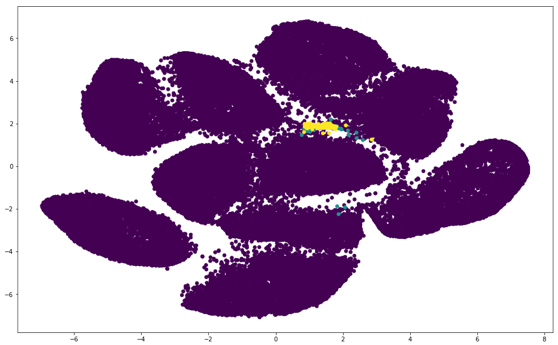}
\caption{2D view of embedding.}
\label{tsne}
\end{subfigure}

\caption{Testing hypothesis.}
\label{hypothesis}
\end{figure}

In our previous experiments that involve out of sample images, we observe that using logit-based uncertainty, we still have images that show low to medium uncertainty. For example, logit-based uncertainties of the model trained on Cifar10 that tested on Cifar100 show many small values. We hypothesize that it is because some out of sample images are intrinsically similar to the training images. To test the hypothesis, we train two models. The first model is trained on the MNIST data set, and using the model’s uncertainty output, we create two subsets of FashionMNIST images of low uncertainty and high uncertainty, with the low uncertainty set having uncertainty values below $0.2$, and the high uncertainty set having uncertainty values above $0.8$. Then we trained another special model that is independent of the first model on both MNIST and FashionMNIST data set. This special model has embedding layers that are shared during training for both data sets, but has separate fully connected layers for MNIST and for FashionMNIST. Then we compare the output of the shared embedding layers for images from the low and high uncertainty sets to the training MNIST images. To compare the output, we first use PCA to reduce the dimension of the output from the shared embedding layer to 10. Then we compute the minimum L2 distance from each low and high uncertainty image to the training MNIST images. The resulting distributions of distances are shown in Figure \ref{min_dist}. The low uncertainty images are more similar to the training images than the high uncertainty images to the training images.

To better visualize the results, we conduct another experiment using the output of the shared embedding layer. We first reduce the dimension of the output to 300 using PCA for computational efficiency, before further decrease the dimension of the output to two with t-SNE, and we plot the resulting samples on the $x$-$y$ plane as shown in Figure \ref{tsne}. The purple dots are the training set of MNIST, the yellow dots represent high uncertainty images, and the cyan dots represent low uncertainty images. As shown in the plot, the low uncertainty images are closer to the training set compare to the high uncertainty images.

The above experiments give us the confidence to claim that some out of sample images are intrinsically similar to the training images with regular convolution layers, thus it is expected that during classification they have relatively low uncertainty values.

\subsection{Simulated Applications}

We propose the following uses of logit-based uncertainty. In classification tasks, we propose to keep the classification result if the uncertainty value associated with the classification model is below a certain threshold $t$, and we send the task to a human expert for confirmation if the uncertainty value is above the threshold. We further assume that human experts have perfect accuracy and there is cost for each manually labeled sample, and a cost for each misclassification error the model makes. The total cost is the number of manually labeled samples plus cost times the number of misclassifications.

\begin{figure}[h]
  \centering
  \includegraphics[height=4.5cm]{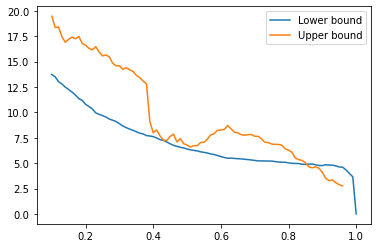}
  \caption{Lower and upper bound of $c$ versus threshold $t$.}
  \label{cost}
\end{figure}

In the following test, we train a model on MNIST, and test on the enlarged USPS images. All five models achieve similar accuracy around $86\%$. Then for each threshold $t$, we find a lower and upper bound on the cost $c$ such that the total cost of using logit-based uncertainty is lower compared to other methods. The lower bound is typically imposed by the ensemble and dropout methods, and the upper bound is typically imposed by the DkNN and evidential methods. Figure \ref{cost} shows the lower and upper bound. When threshold is about $0.85$ to $0.95$, the upper bound is below the lower bound, implying there is no $c$ that makes logit-based model's total cost minimum. However, the bounds are still meaningful if we only compare the logit-based uncertainty to the models that impose the bounds. By chance, the model using logit-based uncertainty has a higher accuracy, therefore, the upper bound for threshold greater than about $0.95$ is infinity and the lower bound goes to 0.  
From the figure, we notice that for almost all threshold, there exist $c$ that makes logit-based uncertainty superior to all other methods tested. Note that compare to DkNN and evidential models, logit-based uncertainty achieves lower total cost if the cost of a mistake is lower, again, showcasing the over-conservative nature of DkNN and evidential models.

\begin{figure}[h]
  \centering
  \includegraphics[height=4.5cm]{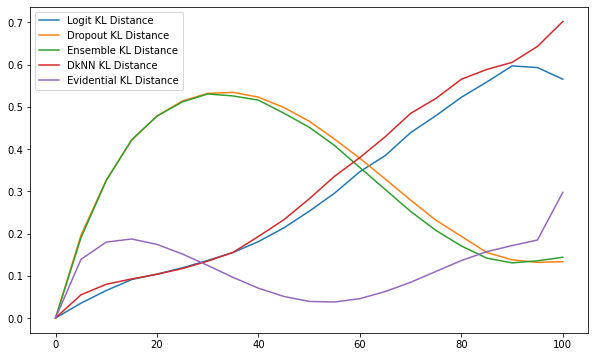}
  \caption{FashionMNIST percentage versus KL distance.}
  \label{KL}
\end{figure}

Another use of logit-based uncertainty involves tests on context drift. We first train a classification model on MNIST, and compute uncertainty on the 10,000 MNIST test set images. We use Gaussian distributions to estimate the distribution of the resulting uncertainty values, and compare it to the distribution of uncertainty values obtained on 10,000 images that are formed from a mixture of the MNIST test set and the FashionMNIST training set. We then compute the KL distance of the two uncertainty distributions obtained, and repeat the process with different portions of the FashionMNIST images in the mixture. Figure \ref{KL} shows the result with the $x$-axis as the percentage in the mixture that are FashionMNIST images, and the $y$-axis   the KL distance of the uncertainty distribution between the mixture of images and the MNIST test set. We expect a good uncertainty measure to produce an increasing KL-distance as portions of FashionMNIST increases, and both DkNN and logit-based uncertainty meet this expectation.

\section{Conclusion}

This paper proposes an uncertainty measure with a novel theoretical foundation for classifications with models of logit outputs. The logit-based uncertainty measure is general in the sense that it is agnostic to the network architecture, to the learning procedure, and to the training task. The extensive experiments show our method’s superior performance on different tasks with different architectures compared to existing uncertainty methods. The innovative GAN tests demonstrate that the intermediate uncertainty values and their order relationships are meaningful, which shows a significant improvement from existing methods. We also demonstrate the potential of the logit-based uncertainty measure in two applications. 

\section*{Acknowledgement}
The majority of this work is completed on Google Cloud Platform and we thank the Google Cloud Platform research credit program for the research credits.

\section*{References}
\small

[1]	Hugo Touvron, Andrea Vedaldi, Matthijs Douze, and Herve Jegou. Fixing the train-test resolution discrepancy. \emph{arXiv preprint arXiv 1906.06423}, 2019

[2]	Heinrich Jiang, Been Kim, Melody Y. Guan, and Maya Gupta. To trust or not to trust a classifier. \emph{NIPS}, 2018 

[3]	Balaji Lakshminarayanan, Alexander Pritzel, and Charles Blundell. Simple and scalable predictive uncertainty estimation using deep ensembles. \emph{NIPS}, 2017.

[4]	Yarin Gal and Zoubin Ghahramani. Dropout as a Bayesian approximation: Representing model uncertainty in deep learning. \emph{ICML, pages 1050–1059}, 2016.

[5]	Murat Sensoy, Lance Kaplan, and Melih Kandemir. Evidential deep learning to quantify classification uncertainty. \emph{NIPS}, 2018

[6]	Charles Blundell, Julien Cornebise, Koray Kavukcuoglu, and Daan Wierstra. Weight uncertainty in neural networks. \emph{ICML}, 2015

[7]	Ian J. Goodfellow, Jean Pouget-Abadie, Mehdi Mirza, Bing Xu, David Warde-Farley, Sherjil Ozair, Aaron Courville, and Yoshua Bengio. Generative adversarial nets. \emph{NIPS}, 2014

[8]	Kimin Lee, Honglak Lee, Kibok Lee, and Jinwoo Shin. Training confidence-calibrated classifiers for detecting out-of-distribution samples. \emph{ICLR}, 2018

[9]	Gao Huang, Zhuang Liu, Laurens van der Maaten, and Kilian Q. Weinberger. Densely connected convolutional networks. \emph{arXiv preprint arXiv 1608.06993v5}, 2018

[10] Rob J. Hyndman. Computing and graphing high density regions. \emph{The American Statistician. Vol. 50, No. 2. pp. 120-126}, 1996

[11] Nicolas Papernot and Patrick McDaniel. Deep k-Nearest neighbors: Towards confident interpretable and robust deep learning. \emph{arXiv preprint arXiv 1803.04765v1}, 2018

[12] Eric J. Hartman, James D. Keeler, and Jacek M. Kowalski. Layered neural networks with Gaussian hidden units as universal approximations. \emph{Neural Computation 2, pp. 210-215}, 1990

[13] Alexander G. de G. Matthews, Jiri Hron, Mark Rowland, Richard E. Turner, and Zoubin Ghahramani. Gaussian process behaviour in wide deep neural networks. \emph{ICLR}, 2018

[14] Jaehoon Lee, Yasaman Bahri, Roman Novak, Samuel S. Schoenholz, Jeffrey Pennington, Jascha Sohl-Dickstein. Deep neural networks as Gaussian processes. \emph{ICLR}, 2018

[15] Radford M. Neal. Priors for infinite networks (tech. rep. no. crg-tr-94-1). \emph{University of Toronto}, 1994

[16] Yiming Xu and Diego Klabjan. Concept drift and covariate shift detection ensemble with lagged labels. \emph{arXiv preprint  arXiv 2012.04759v3}, 2020

\newpage
\section*{Appendix}
\subsection*{Proof of Theorem 1}
Let $G(c)=\sum_{i=1}^{c}\pi_i\mathcal{N}_i$, where $\mathcal{N}_i$ is Gaussian. Consider $c$ neural networks that share the same trainable parameters but the bias of the final layer. The bias of the final layer of $i$-th neural network are normally distributed as $\mathcal{N}_i$. For each of the $c$ neural networks, Theorem 4 from Alexander et al. [13] applies. The output of each of the neural networks converges in distribution to a Gaussian Process. Since we have a finite set of inputs, an immediate result that follows is the outputs of the neural network are jointly Gaussian based on  $\mathcal{N}_{i}^*$. Finally, consider the neural network with bias of the final layer based on distribution $G(c)$. This neural network outputs according to distribution $\mathcal{N}_{i}^*$ with probability $\pi_i$, therefore, the output of the neural network with Gaussian Mixture final layer bias is Gaussian Mixture distributed.
\\\null\hfill $\blacksquare$

\subsection*{Proof of Proposition 2}
The $q_1$-HDR of a Gaussian mixture for class $i$ is the subset $R(f_{1-q_1})$ of the sample space of $X$ defined by $R(f_{1-q_1}) = \{x: gmm_i(x) \geq f_{1-q_1}\}$, where $f_{q_1}$ is the largest constant such that $P(X \in R(f_{1-q_1})) \geq q_1$. Therefore, for every $x$ we have $P(gmm_i(x) < f_{1-q_1}) < 1 - q_1$ and thus $P(\ln(gmm_i(x)) < \ln(f_{1-q_1})) < 1 - q_1$ and $P(s_i(x) < \ln(\max_t(gmm_i(t)/f_{1-q_1}))) \geq q_1$. In turn $\ln(\max_t(gmm_i(t)/f_{1-q_1})) = s_{iq_1}$. Since for every $x \in R(f_{1-q_1})$ we have $gmm_i(x) \geq f_{1-q_1}$, it follows $s_i(x) <  \ln(\max_t(gmm_i(t)/f_{1-q_1}))$ and thus $g_i(s_i(x)) < g_i(\ln(\max_t(gmm_i(t)/f_{1-q_1})))$. Finally, we obtain $g_i(s_i(x)) < g_i(s_{iq_1})$ and $u(x) < u_1$. We can similarly proof the statement for $u_2$.
\\\null\hfill $\blacksquare$
\subsection*{Proof of Proposition 3}
For every $x$ we have $P(x \in R(f_\alpha)) \geq 1 - \alpha$ and $P(\phi(x) \geq f_\alpha) \geq 1 - \alpha$. Using the probability density function of multivariate Gaussian, we obtain $P(r^2(x) < 2 \cdot \ln(f_\alpha \cdot c)) \geq 1 - \alpha$ where $c = (2\pi)^{k/2}det(\Sigma)^{1/2}$, and $k$ is the dimension of the Gaussian. In turn, $P(r(x) < (2 \cdot \ln(f_\alpha \cdot c))^{1/2}) \geq 1 - \alpha$. By definition of cumulative distribution functions, we have $r_\alpha = F^{-1}(1-\alpha) = (2 \cdot \ln(f_\alpha \cdot c))^{1/2}$, and each $x$, $x \in R(f_\alpha)$ implies $\phi(x) \geq f_\alpha$ and thus $r(x) < (2 * \ln(f_\alpha \cdot c))^{1/2}$, which is equivalent to $r(x) < r_\alpha$.
\\\null\hfill $\blacksquare$

\end{document}